\documentclass[10pt,twocolumn,letterpaper]{article}

\usepackage{cvpr}
\usepackage{times}
\usepackage{epsfig}
\usepackage{graphicx}
\usepackage{amsmath}
\usepackage{amssymb}
\usepackage{balance}
\usepackage{multirow}
\usepackage{xspace}
\usepackage{verbatim}
\usepackage{enumerate}
\usepackage[pagebackref=false,breaklinks=true,colorlinks,bookmarks=false,urlcolor=blue,linkcolor=blue,citecolor=blue,pdftitle={Patch-based Probabilistic Image Quality Assessment for Face Selection and Improved Video-based Face Recognition},pdfauthor={Yongkang Wong, Shaokang Chen, Sandra Mau, Conrad Sanderson, Brian C. Lovell}]{hyperref}

\graphicspath{{./}{./figures/}}

\def\Vec#1{{\boldsymbol{#1}}}
\def\Mat#1{{\boldsymbol{#1}}}

\cvprfinalcopy

\def\cvprPaperID{xx}

\begin{document}

\title
	{
	Patch-based Probabilistic Image Quality Assessment for
	\\Face Selection and Improved Video-based Face Recognition
  }

\author
	{
	{Yongkang Wong, Shaokang Chen, Sandra Mau, Conrad Sanderson, Brian C. Lovell}\\
	~\\
  ~~NICTA, PO Box 6020, St Lucia, QLD 4067, Australia \thanks{{\bf Published~in:} IEEE Conference on Computer Vision and Pattern Recognition Workshops (CVPRW), pp.~74--81, 2011.}
  \\
  The University of Queensland, School of ITEE, QLD 4072, Australia%
	}

\maketitle
\thispagestyle{empty}

\maketitle

\begin{abstract}
\vspace{-1ex}

\noindent
In video based face recognition,
face images are typically captured over multiple frames in uncontrolled conditions,
where head pose, illumination, shadowing, motion blur and focus change over the sequence.
Additionally,
inaccuracies in face localisation can also introduce scale and alignment variations.
Using all face images, including images of poor quality,
can actually degrade face recognition performance.
While one solution it to use only the `best' subset of images,
current face selection techniques are incapable of simultaneously handling all of the abovementioned issues.
We propose an efficient patch-based face image quality assessment algorithm
which quantifies the similarity of a face image to a probabilistic face model, representing an `ideal' face.
Image characteristics that affect recognition are taken into account,
including variations in geometric alignment (shift, rotation and scale),
sharpness, head pose and cast shadows.
Experiments on FERET and PIE datasets
show that the proposed algorithm is able to identify images
which are simultaneously the most frontal, aligned, sharp and well illuminated.
Further experiments on a new video surveillance dataset (termed ChokePoint)
show that the proposed method provides better face subsets than existing face selection techniques,
leading to significant improvements in recognition accuracy.

\end{abstract}

~

\section{Introduction}

Video-based identity inference in surveillance conditions
is challenging due to a variety of factors,
including the subjects' motion,
the uncontrolled nature of the subjects,
variable lighting,
and poor quality CCTV video recordings.
This results in issues for face recognition such as low resolution,
blurry images (due to motion or loss of focus),
large pose variations,
and low contrast~\cite{Harandi_IJCV_2009,Sanderson_ICB_2009,Wong_ICPR_2010,facerec-survey04}.
While recent face recognition algorithms can handle faces with moderately challenging illumination conditions~\cite{Harandi_CVPR_2011,Kumar_ICCV_2009,Phillips_PAMI_10,Sanderson_ICB_2009},
strong illumination variations (causing cast shadows~\cite{Sanin_ICPR_2010} and self-shadowing) remain problematic~\cite{shan_amfg_03}.

One approach to overcome the impact of poor quality images
is to assume that such images are outliers in a sequence.
This includes approaches like exemplar extraction using clustering techniques
(eg.~k-means clustering~\cite{Hadid_FGR_2004})
and statistical model approaches for outlier removal~\cite{Berrani_AVSS_2005}.
However, these approaches are not likely to work when most of the images in the sequence have poor quality
--- the good quality images would actually be classified as outliers.

Another approach is explicit subset selection,
where a face quality assessment is automatically made on each image,
either to remove poor quality face images,
or to select a subset comprised of high quality images~\cite{Gao_ICB_2007,Mau_ICVNZ_2010,Nasrollahi_BIOID_2008,Subasic_ISPA_2005}.
This improves recognition performance,
with the additional benefit of reducing the overall computation load during feature extraction and matching~\cite{MOBIO_ICPR_2010}.
The challenge in this approach is finding a good definition for ``face quality''.

Several face image standards have been proposed for face quality assessment
(eg.~ISO/IEC \mbox{19794-5}~\cite{ISO_IEC_19794_5} and ICAO 9303~\cite{ICAO-9303}).
In these standards,
quality can be divided into:
{\bf (i)} {\it image} specific qualities such as sharpness, contrast, compression artifacts,
and
{\bf (ii)} {\it face} specific qualities such as face geometry, pose, eye detectability, illumination angles.

Based in part on the above standards,
many approaches have been proposed to analyse various face and image properties.
For example, face pose estimation using tree structured multiple pose estimators~\cite{Yang_ICPR_2004},
and
face alignment estimation using template matching~\cite{Chang_CIARP_2008}.
Asymmetry analysis has been proposed to simultaneously estimate two qualities:
out-of-plane rotation and non-frontal illumination~\cite{Gao_ICB_2007,ISO-IEC_ICB_2009,Zhang_isvc_2009}.

Since face recognition performance is simultaneously impacted by multiple factors,
being able to detect one or two qualities is insufficient for robust subset selection.
One approach to simultaneously detect multiple quality characteristics
is through a fusion of individual face and image quality measurements.
Nasrollahi and Moeslund~\cite{Nasrollahi_BIOID_2008} proposed a weighted quality fusion approach to combine out-of-plane rotation,
sharpness, brightness, and image resolution qualities.
Rua et al.~\cite{Rua_BIOID_2008} proposed a similar quality assessment approach,
by using asymmetry analysis and two sharpness measurements.
Hsu~et al.~\cite{Hsu_BS_2006} proposed to learn fusion parameters
on multiple quality scores to achieve maximum correlation with matching scores between face pairs.
\mbox{Another} proposed fusion approach uses a Bayesian network to model
the relationships among qualities, image features and matching scores~\cite{Ozay_CVPRW_2009}.
The main drawbacks of the above fusion approaches are:
\begin{itemize}

\item
Fusion-based approaches only perform as well as their individual classifiers.
For example, if a pose estimation algorithm requires accurate facial feature localisation,
the whole fusion framework will fail in the cases where that pose algorithm fails
(such as in low resolution CCTV footage)~\cite{low-res-face-detect01}.

\item
As various properties are measured individually and have different influence on face quality,
it may be difficult to combine them to output a single quality score for the purposes of image selection.

\item
As multiple classifiers as involved,
they are typically more time consuming
and hence may not be suitable for real-time surveillance applications.

\item
Since face matching scores are heavily dependant on system-specific details
(including the input features, matching algorithms and training images), 
quality assessment approaches that learn a fusion model based on match scores
end up being closely tied to the particular system configuration
and hence need to be retrained for each system.

\end{itemize}

\noindent
Simultaneously detecting multiple quality characteristics can also be accomplished
by learning a generic model to define the `ideal' quality.
Luo~\cite{Luo_icip_2004} proposed a learning based approach
where the quality model is trained to correlate with manually labelled quality scores.
However, given the subjective nature of human labelling,
and the fact that humans may not know what characteristics work best
for automatic face recognition algorithms,
this approach may not generate the best quality model for face recognition.

In this paper we propose a straightforward and effective patch-based face quality assessment algorithm,
targeted towards handling images obtained in surveillance conditions.
It quantifies the similarity of a given face to a probabilistic face model,
representing an `ideal' face,
via patch-based local analysis.
Without resorting to fusion,
the proposed algorithm outputs a single score for each image,
with the score simultaneously reflecting the degree of
alignment errors, 
pose variations,
shadowing,
and image sharpness (underlying resolution).
Localisation of facial features (ie.~eyes, nose, mouth) is not required.

We continue the paper as follows.
In Section~\ref{sec:proposed_quality_score} we describe the proposed quality assessment algorithm.
Still image and video datasets used in the experiments are briefly described in in Section~\ref{sec:database}.
Extensive performance comparisons against existing techniques are given in Section~\ref{sec:experiment_quality} (on~still images)
and Section~\ref{sec:experiment_face_rec} (on surveillance videos).
The~main findings are discussed in Section~\ref{sec:conclusions}.

\section{Probabilistic Face Quality Assessment}
\label{sec:proposed_quality_score}

The proposed algorithm is comprised of five steps:
{\bf (1)}~pixel-based image normalisation,
{\bf (2)}~patch extraction and normalisation,
{\bf (3)}~feature extraction from each patch,
{\bf (4)}~local probability calculation,
{\bf (5)}~overall quality score generation via integration of local probabilities.
These steps are elaborated below:

\begin{enumerate}

\item

For a given image {\small $\Mat{I}$},
we perform non-linear pre-processing (log transform) to reduce the dynamic range of data.
Following~\cite{chen_TSMC_2006},
the normalised image {\small $\Mat{I}_{\mathsf{log}}$} is calculated using:
\begin{small}
\begin{equation}
  \Mat{I}_{\mathsf{log}}(r,c) = \ln [\Mat{I}(r,c) + 1]
  \label{equ:log-norm}
\end{equation}%
\end{small}%
where {\small $I(r,c)$} is the pixel intensity located at {\small $(r,c)$}.
Logarithm normalisation amplifies low intensity pixels and compresses high intensity pixels.
This property is helpful in reducing the intensity differences between skin tones.

\item
The transformed image $\Mat{I}_{\mathsf{log}}$
is divided into {\small $N$} overlapping blocks (patches).
Each block {\small $\Mat{b}_i$} has a size of {\small $n \times n$} pixels
and overlap neighbouring blocks by {\small $t$} pixels.
To accommodate for contrast variations between face images,
each patch is normalised to have zero mean and unit variance~\cite{lighting_normalization}.

\item
From each block, a 2D Discrete Cosine Transform (DCT) feature vector is extracted~\cite{Gonzales_2007}.
Excluding the \mbox{0-th} DCT component (as it has no information due to the previous normalisation),
the top {\small $d$} low frequency components are retained.
The low frequency components retain generic facial textures~\cite{modular_PCA},
while largely omitting person-specific information.
At the same time, cast shadows~\cite{lighting_normalization} as well as variations in pose and alignment
can alter the local textures.

\item
For each block location {\small $i$},
the probability of the corresponding feature vector {\small $\Vec{x}_i$} is calculated
using a {\it location specific} probabilistic model:
\begin{small}
\begin{eqnarray}
 \label{equ:probability}
	p(\Vec{x_i} | \Vec{\mu}_i, \Mat{\Sigma}_i)	= \frac
    {
    \exp \left[ -\frac{1}{2}\left(  \Vec{x}_i - \Vec{\mu}_i \right)^T \Mat{\Sigma}_i^{-1} \left(
    \Vec{x}_i - \Vec{\mu}_i \right) \right] }
    {
    \left( 2 \pi \right)^\frac{d}{2} | \Mat{\Sigma}_i | ^\frac{1}{2}
    }
\end{eqnarray}%
\end{small}%
where {\small $\Vec{\mu_i}$} and {\small $\Mat{\Sigma_i}$} are the mean and covariance matrix of a normal distribution.
The model for each location is trained using a pool of frontal faces
with frontal illumination and neutral expression.
All of the training face images are first scaled and aligned to a fixed size,
with each eye located at a fixed location.
We emphasise that during testing, the faces do not need to be aligned.

\item
By assuming that the model for each location is independent,
an overall probabilistic quality score {\small $Q$} for image {\small $\Mat{I}$},
comprised of {\small $N$} blocks, is calculated using:
\begin{small}
\begin{equation}
\label{equ:quality}
    Q(\Mat{I}) = \sum\nolimits_{i=1}^{N} \log p(\Vec{x}_i | \Vec{\mu}_i, \Mat{\Sigma}_i )
\end{equation}%
\end{small}%

\end{enumerate}

The resulting quality score represents the probabilistic similarity of a given face
to an ``ideal'' face (as represented by a set of training images).
A higher quality score reflects better image quality.

\section{Face Datasets}
\label{sec:database}

In this section, we briefly describe the FERET, PIE and ChokePoint face datasets,
as well as their setup for our experiments.

FERET~\cite{Phillips_PAMI_2000} and PIE~\cite{Sim_PAMI_2003}
are used to analyse how accurate the proposed quality assessment algorithm is
for correctly selecting best quality images with several desired characteristics,
compared to other existing methods.
In total, there are 1124 unique subjects in the training phase and 1263 subjects in the test phase.

The ChokePoint dataset contains surveillance videos.
It is used to study the improvement in verification performance gained from subset selection,
using the proposed quality method as well as other approaches.

\vspace{1ex}
\subsection{Setup of Still Image Datasets: FERET and PIE}
\label{sec:database_still}

To study the performance of the proposed method
in terms of correctly selecting images with desired characteristics,
we simulated blurring as well as four alignment errors using images from the `fb' subset of FERET.
Experiments with pose variations (out-of-plane rotation) used dedicated subsets from FERET and PIE.
Experiments with cast shadows used the illumination subset of PIE.

\begin{figure}[!tb]
  \begin{minipage}{1.0\columnwidth}
    \begin{minipage}{0.15\columnwidth}
      \centerline{\includegraphics[width=\columnwidth]{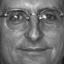}}
      \vspace{-0.8ex}
      \centerline{\footnotesize Aligned}
      \vspace{-0.8ex}
      \centerline{\footnotesize }
    \end{minipage}
    \hfill
    \begin{minipage}{0.15\columnwidth}
      \centerline{\includegraphics[width=\columnwidth]{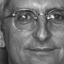}}
      \vspace{-0.8ex}
      \centerline{\footnotesize Horizontal}
      \vspace{-0.8ex}
      \centerline{\footnotesize Shift}
    \end{minipage}
    \hfill
    \begin{minipage}{0.15\columnwidth}
      \centerline{\includegraphics[width=\columnwidth]{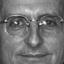}}
      \vspace{-0.8ex}
      \centerline{\footnotesize Vertical}
      \vspace{-0.8ex}
      \centerline{\footnotesize Shift}
    \end{minipage}
    \hfill
    \begin{minipage}{0.15\columnwidth}
      \centerline{\includegraphics[width=\columnwidth]{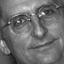}}
      \vspace{-0.8ex}
      \centerline{\footnotesize In-Plane}
      \vspace{-0.8ex}
      \centerline{\footnotesize Rotation}
    \end{minipage}
    \hfill
    \begin{minipage}{0.15\columnwidth}
      \centerline{\includegraphics[width=\columnwidth]{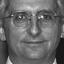}}
      \vspace{-0.8ex}
      \centerline{\footnotesize Scale}
      \vspace{-0.8ex}
      \centerline{\footnotesize Change}
    \end{minipage}
    \hfill
    \begin{minipage}{0.15\columnwidth}
      \centerline{\includegraphics[width=\columnwidth]{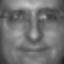}}
      \vspace{-0.8ex}
      \centerline{\footnotesize Blurring}
      \vspace{-0.8ex}
      \centerline{\footnotesize ~}
    \end{minipage}
  \end{minipage}
  
  ~
  
  \caption
    {
    Examples of simulated image variations on FERET.
    }
  \label{fig:feret-sample}
\end{figure}

The generated alignment errors%
\footnote{The generated alignment errors are representatives of real-life characteristics of automatic face localisation/detection algorithms~\cite{Rodriguez_IVC_2006}.}
are:
horizontal shift and vertical shift
(using displacements of
{\small $0$}, {\small $\pm2$}, {\small $\pm4$}, {\small $\pm6$}, {\small $\pm8$} pixels),
in-plane rotation
(using rotations of {\small $0^{\circ}$}, {\small $\pm10^{\circ}$}, {\small $\pm20^{\circ}$}, {\small $\pm30^{\circ}$}),
and scale variations (using scaling factors of
{\small $0.7$}, {\small $0.8$}, {\small $0.9$}, {\small $1.0$}, {\small $1.1$}, {\small $1.2$}, {\small $1.3$)}.
For sharpness variations,
each original image is first downscaled to three sizes
({\small $48\times48$}, {\small $32\times32$} and {\small $16\times16$} pixels)
then rescaled to the baseline size of {\small $64\times64$} pixels.
See Fig.~\ref{fig:feret-sample} for examples.

FERET provides the dedicated `b' subset with pose variations,
containing
out-of plane rotations of {\small $0^{\circ}$}, {\small $\pm15^{\circ}$}, {\small $\pm25^{\circ}$}, {\small $\pm40^{\circ}$}, {\small $\pm60^{\circ}$}.
PIE also provides a dedicated subset with pose variations,
though with a smaller set of rotations
({\small $0^{\circ}$}, {\small $\pm22.5^{\circ}$}, {\small $\pm45^{\circ}$}, {\small $\pm67.5^{\circ}$}).

\begin{figure}[!tb]
  \centering
  \begin{minipage}{1.0\columnwidth}
    \begin{minipage}{0.15\columnwidth}
      \centerline{\includegraphics[width=\columnwidth]{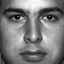}}
      \vspace{-0.8ex}
      \centerline{\footnotesize Subset 1}
      \vspace{-0.8ex}
      \centerline{\footnotesize ($0^\circ$)}
    \end{minipage}
    \hfill
    \begin{minipage}{0.15\columnwidth}
      \centerline{\includegraphics[width=\columnwidth]{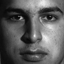}}
      \vspace{-0.8ex}
      \centerline{\footnotesize Subset 2}
      \vspace{-0.8ex}
      \centerline{\footnotesize ($16^\circ$-~$21^\circ$)}
    \end{minipage}
    \hfill
    \begin{minipage}{0.15\columnwidth}
      \centerline{\includegraphics[width=\columnwidth]{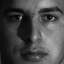}}
      \vspace{-0.8ex}
      \centerline{\footnotesize Subset 3}
      \vspace{-0.8ex}
      \centerline{\footnotesize ($31^\circ$-~$32^\circ$)}
    \end{minipage}
    \hfill
    \begin{minipage}{0.15\columnwidth}
      \centerline{\includegraphics[width=\columnwidth]{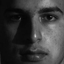}}
      \vspace{-0.8ex}
      \centerline{\footnotesize Subset 4}
      \vspace{-0.8ex}
      \centerline{\footnotesize ($37^\circ$-~$38^\circ$)}
    \end{minipage}
    \hfill
    \begin{minipage}{0.15\columnwidth}
      \centerline{\includegraphics[width=\columnwidth]{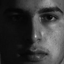}}
      \vspace{-0.8ex}
      \centerline{\footnotesize Subset 5}
      \vspace{-0.8ex}
      \centerline{\footnotesize ($44^\circ$-~$47^\circ$)}
    \end{minipage}
    \hfill
    \begin{minipage}{0.15\columnwidth}
      \centerline{\includegraphics[width=\columnwidth]{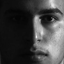}}
      \vspace{-0.8ex}
      \centerline{\footnotesize Subset 6}
      \vspace{-0.8ex}
      \centerline{\footnotesize ($54^\circ$-~$67^\circ$)}
    \end{minipage}
  \end{minipage}
  
  ~
  
  \caption
    {
    Examples from PIE with strong directed illumination, causing self-shadowing.
    }
  \label{fig:pie_sample}
  \vspace{-1ex}
\end{figure}

The illumination subset of PIE was used to assess performance in various cast shadow conditions.
In our experiments, 
we divided the frontal view images into six subsets\footnote {
Subset~1: light source 8, 11, 20;
Subset~2: light source 6, 7, 9, 12, 19, 21;
Subset~3: light source 5, 10, 13, 14;
Subset~4: light source 18, 22;
Subset~5: light source 4, 15;
Subset~6: light source 2, 3, 16, 17.
}
based on the angle of the corresponding light source.
Subset 1 has the most frontal light sources,
while subset 6 has the largest light sources angle ({\small $54^\circ$-~$67^\circ$}). 
See Fig.~\ref{fig:pie_sample} for examples.

\subsection{Surveillance Videos: ChokePoint Dataset}
\label{sec:database_video}

\begin{figure}[!tb]
\centering
\begin{tabular}{c|c|c|c}
  Camera Rig & Camera~1 & Camera~2 & Camera~3
  \\
  \hspace{-1.4ex}
  \includegraphics[width=0.3\columnwidth]{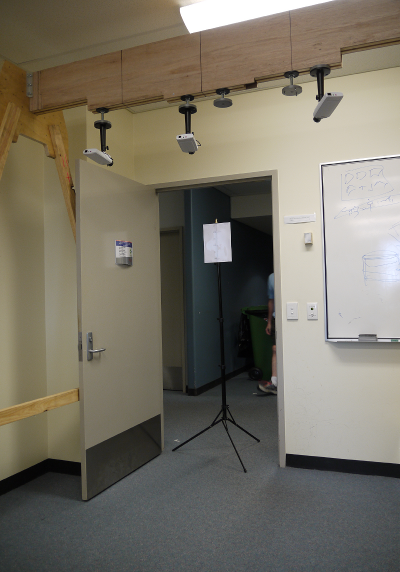}
  \hspace{-1.4ex}
  &
  \hspace{-1.4ex}
  \includegraphics[width=0.21\columnwidth]{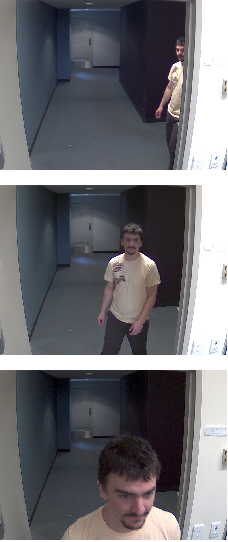}
  \hspace{-1.4ex}
  &
  \hspace{-1.4ex}
  \includegraphics[width=0.21\columnwidth]{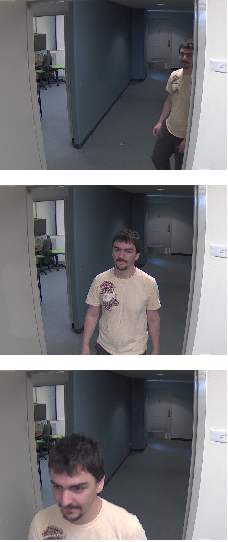}
  \hspace{-1.4ex}
  &
  \hspace{-1.4ex}
  \includegraphics[width=0.21\columnwidth]{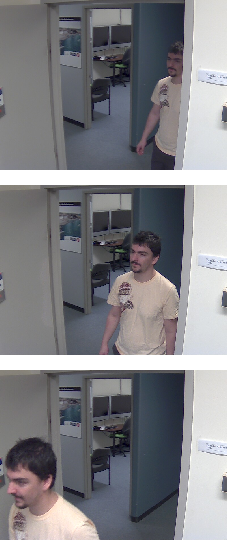}
  \hspace{-1.4ex}
  \\
  \end{tabular}
  ~
  
  \caption
    {
    An example of the recording setup used for the {\it ChokePoint} dataset.
    A camera rig contains 3~cameras placed just above a door,
    used for simultaneously recording the entry of a person from 3~viewpoints.
    The variations between viewpoints allow for variations in walking directions,
    facilitating the capture of a near-frontal face by one of the cameras.
    }
  \label{fig:chokept_rig}
  
  ~
  
  ~
  
  ~
\end{figure}

\begin{figure}[!tb]
  \centering
  
  \begin{minipage}{1.0\columnwidth}
    \begin{minipage}{0.49\columnwidth}
      \centerline{\includegraphics[width=1.0\columnwidth]{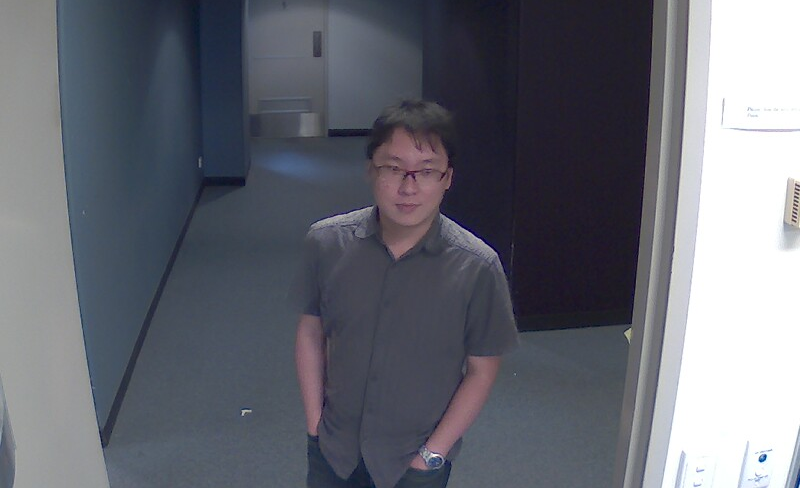}}
    \end{minipage}
    \hfill
    \begin{minipage}{0.49\columnwidth}
      \centerline{\includegraphics[width=1.0\columnwidth]{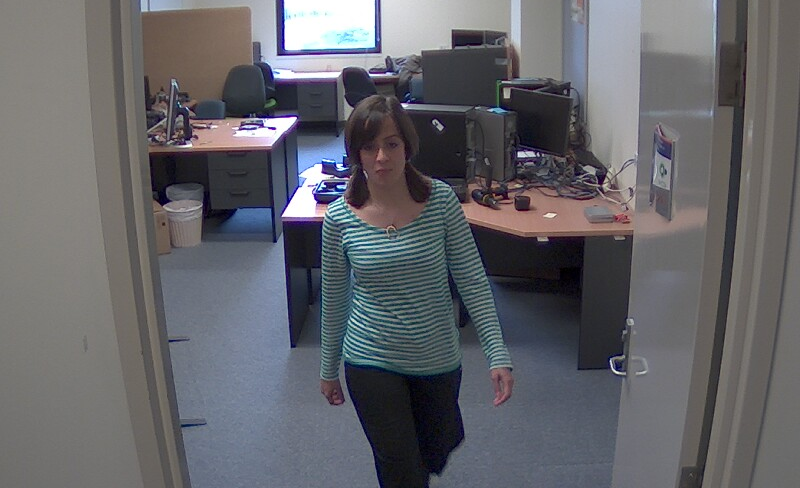}}
    \end{minipage}
  \end{minipage}
  
  \vspace{1ex}
  
  \begin{minipage}{1.0\columnwidth}
    \begin{minipage}{0.49\columnwidth}
      \centerline{\includegraphics[width=1.0\columnwidth]{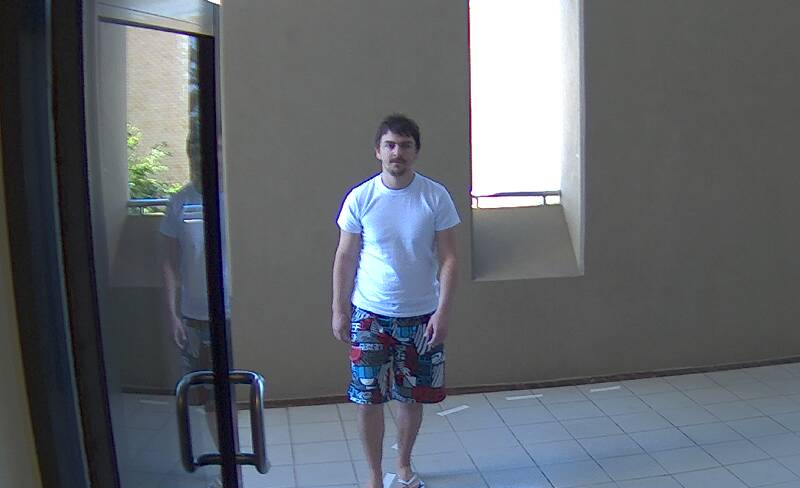}}
    \end{minipage}
    \hfill
    \begin{minipage}{0.49\columnwidth}
      \centerline{\includegraphics[width=1.0\columnwidth]{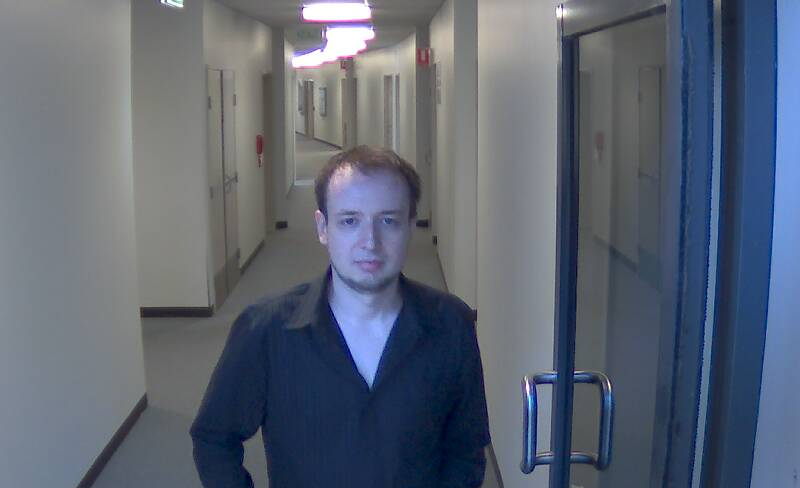}}
    \end{minipage}
  \end{minipage}
  
  \vspace{1.5ex}
  
  \caption
    {
    Example shots from the ChokePoint dataset,
    showing portals with various backgrounds.
    }
  \label{fig:example_chokePt}
  
  ~
\end{figure}

We collected a new video dataset\footnote{\href{http://arma.sourceforge.net/chokepoint/}{\tt http://arma.sourceforge.net/chokepoint/}},
termed {\it ChokePoint},
designed for experiments in person identification/verification 
under real-world surveillance conditions using existing technologies. 
An array of three cameras was placed above several portals (natural choke points in terms of pedestrian traffic)
to capture subjects walking through each portal in a natural way (see Figs.~\ref{fig:chokept_rig} and~\ref{fig:example_chokePt}). 

While a person is walking through a portal,
a sequence of face images (ie.~a face set) can be captured.
Faces in such sets will have variations in terms of illumination conditions, pose, sharpness,
as well as misalignment due to automatic face localisation/detection~\cite{Rodriguez_IVC_2006,Sanderson_ICB_2009}.
Due to the three camera configuration,
one of the cameras is likely to capture a face set where a subset of the faces is near-frontal.

The dataset consists of 25 subjects (19 male and 6 female) in portal 1
and 29 subjects (23 male and 6 female) in portal 2. 
In total, it consists of 48 video sequences and 64,204 face images.
Each sequence was named according to the recording conditions (eg.~P2E\_S1\_C3)
where P, S, and C stand for {\it portal}, {\it sequence} and {\it camera}, respectively.
E~and~L indicate subjects either {\it entering} or {\it leaving} the portal.
The numbers indicate the respective portal, sequence and camera label.
For example,
P2L\_S1\_C3 indicates that the recording was done in Portal 2,
with people leaving the portal,
and captured by camera 3 in the first recorded sequence.

In this paper, all the experiments were performed with the video-to-video verification protocol.
In this protocol, video sequences are divided into two groups ({\small $G1$} and {\small $G2$}),
where each group played the role of development set and evaluation set in turn.
Parameters can be first learned on the development set and then applied on the evaluation set.
The average verification rate is used for reporting results.
In our experiments we selected the frontal view cameras (shown in Table~\ref{tab:chokePt_protocol}).
In each group, each sequence takes turn to be the gallery,
with the the leftover sequences becoming the probe.

\begin{table} [!b]
  \caption
    {
    ChokePoint video-to-video verification protocol.
    Sequences are divided into two groups (G1 and G2).
    Listed sequences contain faces with the most frontal pose view.
    P, S, and C stand for {\it portal}, {\it sequence} and {\it camera}, respectively.
    E~and~L indicate subjects {\it entering} or {\it leaving} the portal.
    The numbers indicate the respective portal, sequence and camera label.
    For example,
    P2L\_S1\_C3 indicates that the recording was done in Portal 2,
    with people leaving the portal,
    and captured by camera 3 in the first recorded sequence.
    }
  \label{tab:chokePt_protocol}
  \begin{small}
  \begin{center}
    \begin{tabular}{|c|c c c c|}
      \hline
                          &              &             &             &             \\ [-9.0px]
      \multirow{2}{*}{G1} &  P1E\_S1\_C1 & P1E\_S2\_C2 & P2E\_S2\_C2 & P2E\_S1\_C3 \\
                          &  P1L\_S1\_C1 & P1L\_S2\_C2 & P2L\_S2\_C2 & P2L\_S1\_C1 \\
      \hline
                          &              &             &             &             \\ [-9.0px]
      \multirow{2}{*}{G2} &  P1E\_S3\_C3 & P1E\_S4\_C1 & P2E\_S4\_C2 & P2E\_S3\_C1 \\
                          &  P1L\_S3\_C3 & P1L\_S4\_C1 & P2L\_S4\_C2 & P2L\_S3\_C3 \\
      \hline
    \end{tabular}
  \end{center}
  \end{small}
\end{table}

\section{Experiments on Still Images}
\label{sec:experiment_quality}

In this section, we evaluate how well the proposed quality assessment method can identify the best quality faces
when presented with both good and poor quality faces.
The proposed method was compared with:
{\bf (i)}~a~score fusion method using pixel based asymmetry analysis and two sharpness analyses (denoted as {\it Asym\_shrp})~\cite{Rua_BIOID_2008},
{\bf (ii)}~asymmetry analysis with Gabor features (denoted as {\it Gabor\_asym})~\cite{ISO-IEC_ICB_2009},
{\bf (iii)}~the classical Distance From Face Space (DFFS) method~\cite{Bae_2005}.

The `fa' subset of FERET,
containing frontal faces with frontal illumination and neutral expression,
was used to train the location specific probabilistic models in the proposed method.
The `fa' subset was also used to select the decision threshold for rejecting ``poor'' quality images.
The `fa' subset was not used for any other purposes.

Based on preliminary experiments,
closely cropped face images were scaled to {\small $64\times64$} pixels,
the block size was set to {\small $8\times8$} pixels,
with a {\small $7$} pixels overlap of neighbouring blocks.
The preliminary experiments also suggested that using just 3 DCT coefficients was sufficient.
This configuration was used in all experiments.
The quality assessment methods were implemented with the aid of the Armadillo C++ library~\cite{Armadillo_2010}.

\begin{table*}
\centering
  \caption
    {
    Quality assessment of alignment errors and sharpness variations on FERET `fb' and
    all six PIE illumination subsets.
    Each value in the table indicates the percentage of the best aligned image in each variation type
    being assigned to have the highest quality score.
    For example, out of the set of faces with rotations of
    {\small $0^{\circ}$}, {\small $\pm10^{\circ}$}, {\small $\pm20^{\circ}$}, {\small $\pm30^{\circ}$},
    the value indicates the percentage of {\small $0^{\circ}$} faces labelled as ``high'' quality.
    The variations included:
    horizontal shift (HS),
    vertical shift (VS),
    in-plane rotation (RT),
    scale (SC),
    sharpness (SH).
    The `overall' columns indicate the average performance of the above variations.
    Best performance is highlighted in bold.
    }
  \label{tab:geometry-dataset}

  \vspace{1.5ex}

  \begin{small}
  \begin{tabular}{|l|c|c|c|c|c|c||c|c|c|c|c|c|}
  \cline{2-13}
  \multicolumn{1}{c}{} & \multicolumn{6}{|c||}{}          & \multicolumn{6}{|c|}{}                  \\ [-9.0px]
  \multicolumn{1}{c}{} & \multicolumn{6}{|c||}{FERET `fb'} & \multicolumn{6}{|c|}{PIE illumination}   \\  \cline{2-13}
  \multicolumn{1}{c|}{}
  &               &               &               &               &                 &
  &               &               &               &               &                 &                \\ [-9.0px]
  \multicolumn{1}{c|}{}
  & HS            & VS            & RT            & SC            & SH              & overall
  & HS            & VS            & RT            & SC            & SH              & overall        \\  \hline
  &               &               &               &               &                 &
  &               &               &               &               &                 &                \\ [-9.0px]
  Asym\_shrp~\cite{Rua_BIOID_2008}
  & 44.4          & ~~7.7         &   79.8        &   ~~7.4       & \textbf{100.0}  & 47.9
  & 10.3          & ~~4.0         &   40.4        & ~~2.4         & \textbf{100.0}  & 31.4           \\ \hline
  &               &               &               &               &                 &
  &               &               &               &               &                 &                \\ [-9.0px]
  Gabor\_asym~\cite{ISO-IEC_ICB_2009}
  & 52.1          & ~~3.1         & 93.9          & 11.5          & ~~49.0          & 41.9
  & 24.7          & ~~1.5         & 66.4          & 10.7          & ~~29.0          & 26.5           \\ \hline
  &               &               &               &               &                 &
  &               &               &               &               &                 &                \\ [-9.0px]
  DFFS~\cite{Bae_2005}
  & 75.6          & 71.9          & 98.7          & 62.5          & ~~~~0.7         & 61.9
  & 64.4          & 62.4          & \textbf{99.6} & \textbf{44.4} & ~~~~2.3         & 54.6          \\ \hline
  &               &               &               &               &                 &
  &               &               &               &               &                 &                \\ [-9.0px]
  Proposed
  & \textbf{83.4} & \textbf{85.4} & \textbf{99.6} & \textbf{73.0} & ~~99.8          & \textbf{88.2}
  & \textbf{65.9} & \textbf{62.6} &   98.8        & 37.0          & ~~95.9          & \textbf{72.0} \\ \hline
  \end{tabular}
  \end{small}

  ~

  ~

\end{table*}

\subsection{Quality Assessment of Faces with Variations in Alignment, Scale and Sharpness}

In this experiment we evaluated the efficacy of each method
to detect the best aligned images within a set of images that have a particular image variation.
For example,  out of the set of faces with rotations of
{\small $0^{\circ}$}, {\small $\pm10^{\circ}$}, {\small $\pm20^{\circ}$}, {\small $\pm30^{\circ}$},
we measured the percentage of {\small $0^{\circ}$} faces that were labelled as ``high'' quality.

Results for variations in shift, rotation and scale, shown in Table~\ref{tab:geometry-dataset},
indicate that the proposed method consistently achieved the best or near-best performance across most of the variations.
The results on the six PIE illumination subsets
indicate that even in the presence of cast shadows,
the proposed method can achieve good results,
with the exception of images with scale changes.
Averaging over all variations, the proposed method achieved the best results.

The asymmetry-based analysis methods (Gabor\_asym and Asym\_sharp)
could not reliably detect vertical alignment errors and scale variations.
Gabor\_asym also performed poorly for detecting images with various sharpness variations.
Asym\_shrp addressed this by combining asymmetry analysis with two image sharpness measurements.
Despite that,
the overall performance of Asym\_shrp was still poor.

The performance of DFFS on alignment errors was consistent
but generally lower than the proposed method.
Notably, DFFS failed to detect images with the best sharpness.

\subsection{Quality Assessment on Pose Variations}
\vspace{1ex}

\begin{table*}
  \hrule
  ~
  \centering
  \caption
    {
		\small
		Quality assessment of pose variations on the pose subsets of FERET and PIE.
		Each value in the table indicates the percentage of images with a particular pose angle
		that were assigned to have the highest quality score.
		Best performance is highlighted in bold.
		}
  \label{tab:pose-database}

  \vspace{1.5ex}

  \begin{small}
	\begin{tabular}{|l|c|c|c|c|c|c|c|c|c|}%
  \cline{2-10}
	\multicolumn{1}{c}{}	& \multicolumn{9}{|c|}{} \\ [-8.0px]
	\multicolumn{1}{c}{}	& \multicolumn{9}{|c|}{FERET pose subset} \\ \cline{2-10}
	\multicolumn{1}{c|}{}	&       &       &       &      &        &   	  &     &       &	\\ [-8.0px]
	\multicolumn{1}{c|}{}	& $-60^{\circ}$ & $-40^{\circ}$ & $-25^{\circ}$ & $-15^{\circ}$ & $0^{\circ}$ & $+15^{\circ}$ & $+25^{\circ}$ & $+40^{\circ}$ & $+60^{\circ}$ \\  \hline
												&       &       &       &       &        &    	   &     &     &	   \\ [-8.0px]
	Asym\_shrp~\cite{Rua_BIOID_2008} &  0    &  0    & 0.5   &  30.5 &  68.0  &  1    & 0   &	0
	&  0 \\ \hline &       &       &       &       &        &   	     &     &     &	   \\ [-8.0px]
	Gabor\_asym~\cite{ISO-IEC_ICB_2009}	&  2    &  5.5  & 7.5   &  24.5 &  58.0  &  2.5    & 0   &
	0  &  0 \\ \hline &       &       &       &       &        &    	   &     &     &	   \\ [-8.0px]
	DFFS~\cite{Bae_2005}		&  0    &  0    & 0     &  5    &  \textbf{92.0}    &  3      & 0
	&	0  &  0 \\ \hline &       &       &       &       &        &   	     &     &     &	   \\ [-8.0px]
	Proposed             	&  0    &  0    & 0.5   &  28   &  68.5  &  3      & 0   &	0  &  0 \\ \hline
	\multicolumn{10}{c}{}	\\ [-6.0px]
	\cline{2-10}
  \multicolumn{1}{c}{}	& \multicolumn{9}{|c|}{} \\ [-8.0px]
	\multicolumn{1}{c}{}	& \multicolumn{9}{|c|}{PIE pose subset} \\ \cline{2-10}
	\multicolumn{1}{c|}{}	&       &       &       &      &        &   	  &     &       &	\\ [-8.0px]
	\multicolumn{1}{c|}{}	& $-67.5^{\circ}$ & $-45^{\circ}$ & $-22.5^{\circ}$ & --- & $0^{\circ}$ & --- & $+22.5^{\circ}$ & $+45^{\circ}$ & $+67.5^{\circ}$ \\  \hline
									&       &       &       &       &        &    	   &     &      &	   \\ [-8.0px]
	Asym\_shrp~\cite{Rua_BIOID_2008} &  0    &  0    & 2.94  &  ---  &  \textbf{94.1}  &  ---    &
	1.5 &	1.5 &  0	\\ \hline &       &       &       &       &        &    	   &     &      &	   \\ [-8.0px]
	Gabor\_asym~\cite{ISO-IEC_ICB_2009} &  0    &  8.8  & 10.3  &  ---  &  73.5  &  ---    & 5.9 &
	1.5 &  0 \\ \hline &       &       &       &       &        &    	   &     &      &	   \\ [-8.0px]	
	DFFS~\cite{Bae_2005}  &  0    &  1.5  & 11.8  &  ---  &  79.4  &  ---    & 7.4 &	0   &  0
	\\ \hline &       &       &       &       &        &    	   &     &      &	   \\ [-8.0px]
	Proposed       	&  0    &  0    & 4.4   &  ---  &  91.2  &  ---    & 4.4 &	0   &  0 \\ \hline
  \end{tabular}
  \end{small}

  ~

  ~

\end{table*}

In this experiment we evaluated the ability of each method to detect the most frontal faces
in a set that included frontal and non-frontal (out-of-plane rotated) faces.
The results, shown in Table~\ref{tab:pose-database},
indicate that the proposed method consistently achieves
second best performance on both FERET and PIE,
with its performance on PIE being quite close to the top performer (Asym\_shrp).

We note that on FERET the visual differences
between faces at {\small $0^{\circ}$} and {\small $\pm 15^{\circ}$} are minimal,
which can explain why a significant proportion of faces at {\small $-15^{\circ}$}
was classified as ``frontal'' by the proposed method.

While DFFS gave the best performance on FERET,
its performance dropped on PIE.
As there is an overlap between the subjects in the `fa' and pose subsets in FERET (where `fa' was used for training),
the inconsistency in performance across FERET and PIE suggests
that DFFS might be over trained to the training dataset.

The performance of Asym\_shrp and the proposed method is considerably better on PIE than on FERET.
We conjecture that this is due to the larger pose variation between frontal faces
and faces with the smallest pose angle ({\small $\pm 22.5^{\circ}$}),
in contrast to {\small $\pm 15^{\circ}$} on FERET.

\begin{table}[!b]
  \hrule
  \vspace{1ex}
  \centering
  \caption
    {
    Quality assessment of images with cast shadows from the PIE dataset.
    Each value in the table indicates the percentage of images with a particular illumination direction
    that were assigned to have the highest quality score.
    The illumination ranged from frontal (subset~1)
    to strongly directed (subset~6) where there are strong shadows (see Fig.~\ref{fig:pie_sample}).
    }
  \label{tab:PIE-illum}

  \vspace{1ex}
  
  \begin{small}
  \begin{tabular}{|@{\hspace{3px}}l@{\hspace{3px}}|c|c|c|c|c|c|}%
  \cline{2-7}
  \multicolumn{1}{c|}{}   & \multicolumn{6}{c|}{}       \\ [-9.0px]
  \multicolumn{1}{c|}{}   & \multicolumn{6}{c|}{PIE illumination subset} \\
  \cline{2-7}
  \multicolumn{1}{c|}{}   &              &       &        &        &       &       \\ [-9.0px]
  \multicolumn{1}{c|}{}   &           1  &     2 &     3  &     4  &     5 &     ~~6~~ \\ \hline
                          &              &       &        &        &       &       \\ [-9.0px]
  Asym\_shrp~\cite{Rua_BIOID_2008}
                          &\textbf{97.1} &  ~2.9 &     0  &     0  &     0 &     0 \\ \hline
                          &              &       &        &        &       &       \\ [-9.0px]
  Gabor\_asym~\cite{ISO-IEC_ICB_2009}
                          &        51.5  &  ~5.9 &  ~2.9  &  39.7  &  ~4.4 &     0 \\ \hline
                          &              &       &        &        &       &       \\ [-9.0px]
  DFFS~\cite{Bae_2005}    &           0  &     0 &  ~4.4  &  88.2  &  ~7.4 &     0 \\ \hline
                          &              &       &        &        &       &       \\ [-9.0px]
  Proposed                &        94.1  &  ~5.9 &     0  &     0  &     0 &     0 \\ \hline
  \end{tabular}
  \end{small}
\end{table}

\subsection{Quality Assessment on Cast Shadow Variations}
\label{sec:exp-illum}
\vspace{-0.5ex}

Here we evaluated the accuracy of selecting frontal face images with the least amount amount of cast shadow
within a set of images subject to varying illumination direction.
The direction ranged from frontal (subset~1) to side (subset~6),
where severe cast shadows exist (as shown in Fig.~\ref{fig:pie_sample}).

The results, presented in Table~\ref{tab:PIE-illum},
show that Asym\_shrp achieved the best performance
(correctly labelling frontally illuminated faces as having high quality),
with the proposed method a close second.
In contrast,
Gabor\_asym was confused between subsets~1 and~4,
while DFFS erroneously labelled most faces in subset~4 (containing significant shadows) as having the highest quality.

\section{Experiments on Video: Subset Selection}
\label{sec:experiment_face_rec}
\vspace{-0.5ex}

In this section, 
we study the effectiveness of using quality measurements
to select a subset of images for video-based face verification.
To demonstrate the effectiveness of the quality assessment for a variety of face recognition systems,
we used two facial feature extraction algorithms 
and two classification techniques, 
specifically designed for dealing with sets of faces (ie.~image set matching).

Specifically,
we separately used Multi-Region Histograms (MRH)~\cite{Sanderson_ICB_2009}
and Local Binary Patterns (LBP)~\cite{Ahonen_ECCV_2004} to extract features from each face.
The comparison between two sets of faces was performed using
{\bf (i)}~Mutual Subspace Method~(MSM)~\cite{Yamaguchi_FGR_1998}
(for both MRH and LBP),
and 
\mbox{{\bf (ii)}~feature} averaging~\cite{KANG_EURASIP_2010,Mau_ICVNZ_2010} (for MRH only).

The experiments were conducted on the ChokePoint dataset,
using the video-to-video protocol (see Sec.~\ref{sec:database_video}).
Each set of face images for a particular person
was rank ordered according to the quality scores of the images,
followed by keeping the top {\small $N$} images. 

As per Section~\ref{sec:experiment_quality},
the proposed face quality measurement method was compared against three other methods:
Asym\_shrp, Gabor\_asym and DFFS.
The `fa' subset of FERET, which is totally independent from ChokePoint,
was used for training DFFS and the proposed quality measurement method.

In the first experiment,
{\small $N$} varied from {\small $4$} to {\small $16$}.
The results, reported in Table~\ref{tab:chokePt_SC_msm},
indicate that the proposed quality measurement method 
consistently leads to better face verification performance than the other three methods,
regardless of the facial feature extraction algorithm used.
The improvement is most prevalent for {\small $N=4$},
indicating that the proposed method assigns high scores to high quality images
more accurately.

\begin{table}[!t]
  \centering
  \caption
    {
    Video-based face verification performance on the {\it ChokePoint} dataset,
    using MRH and LBP feature extraction algorithms coupled with 
    the Mutual Subspace Method (MSM) for classifying face sets.
    Each set of face images for a particular person
    was rank ordered according to the quality scores of the images,
    followed by retaining top $N$ quality images (ie.~$N$ is the subset size).
    Faces were segmented using automatic face localisation (detection).
    The~average face verification rate is reported (see Sec.~\ref{sec:database_video}).
    Best performance is highlighted in bold.
    }
    \label{tab:chokePt_SC_msm}

    \vspace{1ex}

    \begin{small}
    \begin{tabular}{| @{\hspace{3px}}l@{\hspace{3px}}         | 
                     c |  c | @{\hspace{4px}}c@{\hspace{4px}} || 
                     c |  c | @{\hspace{4px}}c@{\hspace{4px}} | }
    \cline{2-7}
    \multicolumn{1}{c}{}  & \multicolumn{6}{|c|}{Recognition Method} \\ \hline
    \multicolumn{1}{|c}{Subset Selection}  & \multicolumn{3}{|c||}{MRH + MSM}  & \multicolumn{3}{c|}{LBP + MSM} \\
    \cline{2-7}
    \multicolumn{1}{|c|}{Method} & N=4     & N=8     & N=16   & N=4    & N=8    & N=16   \\ \hline
                          &         &         &        &        &        &        \\ [-9.0px] 
    Asym\_shrp~{\footnotesize \cite{Rua_BIOID_2008}}
                          & 67.5    & 70.3    & 75.4   & 65.3   & 67.6   & 70.5   \\       
                          &         &         &        &        &        &        \\ [-9.0px]
    Gabor\_asym~{\footnotesize \cite{ISO-IEC_ICB_2009}}
                          & 75.4    & 78.6    & 84.0   & 69.3   & 71.4   & 74.5   \\
                          &         &         &        &        &        &        \\ [-9.0px]
    DFFS~{\footnotesize \cite{Bae_2005}}
                          & 74.7    & 78.1    & 83.4   & 69.4   & 70.3   & 74.6   \\
                          &         &         &        &        &        &        \\ [-9.0px]
    Proposed              & \textbf{82.5} & \textbf{84.5} & \textbf{86.7} 
                          & \textbf{73.5} & \textbf{74.7} & \textbf{75.8} \\
    \hline
  \end{tabular}
  \end{small}
  
  ~
\end{table}

In the second experiment, 
{\small $N$} varied from {\small $1$} to the size of the set (labelled as ``all'').
Each face set was represented by an average MRH signature;
LBP feature extraction was not used as it isn't suitable for feature averaging.
Face sets were then compared by using an {\small $L_1$}-norm based distance between
their corresponding average MRH signatures~\cite{KANG_EURASIP_2010,Mau_ICVNZ_2010,Sanderson_ICB_2009}.

\begin{figure}[!tb]
  \centering
  \includegraphics[width=1.0\columnwidth]{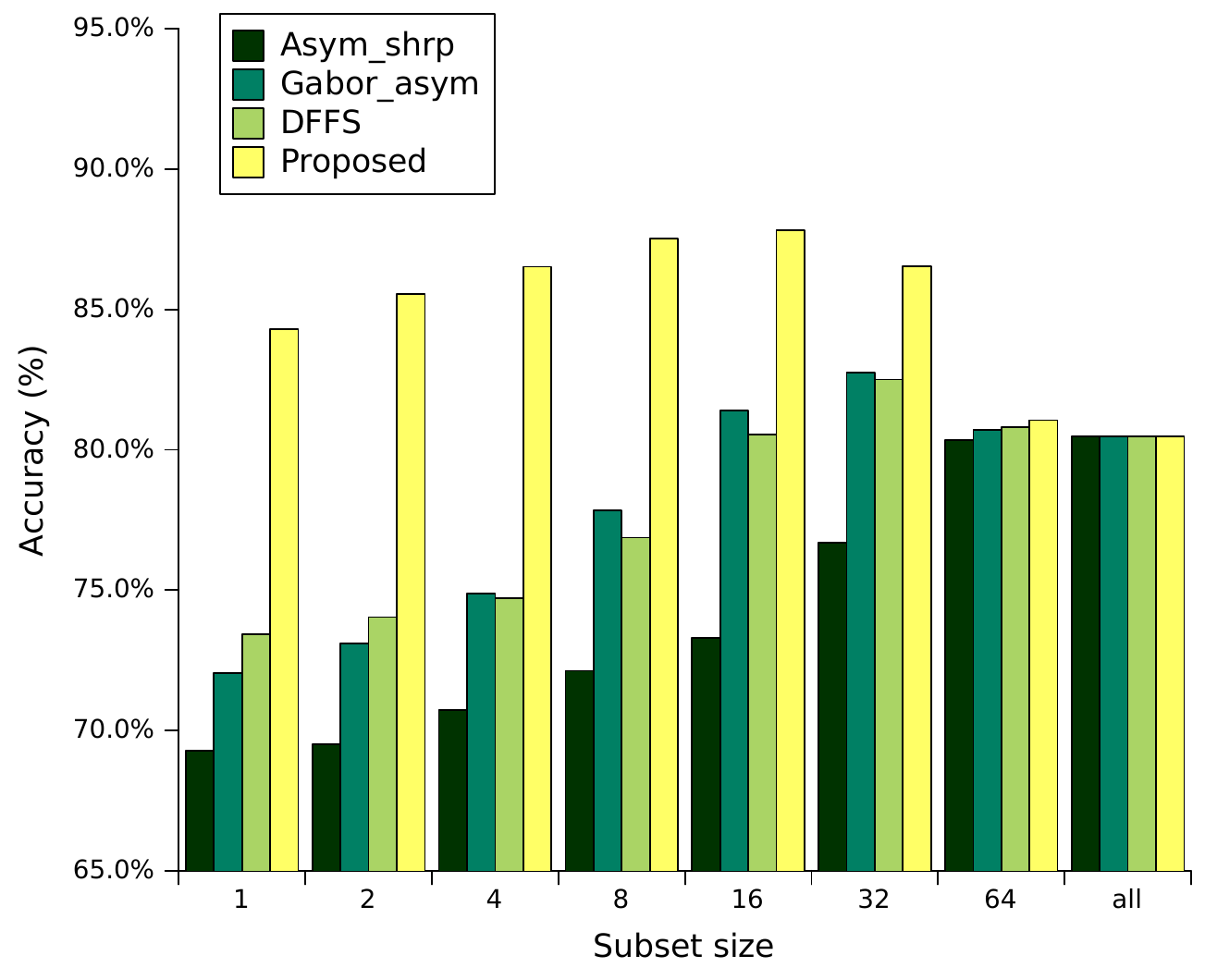}
  \caption
    {
    \small
    Video-based face verification performance on the {\it ChokePoint} dataset using average MRH signatures.
    Each set of face images for a particular person
    was rank ordered according to the quality scores of the images,
    followed by selecting a predefined number of top quality images to create a subset.
    Faces were segmented using automatic face localisation (detection).
    The average face verification rate is reported (see Sec.~\ref{sec:database_video}).
    }
  \label{fig:face_rec_chokePt}
  ~
  \hrule
\end{figure}

From the results shown in Fig.~\ref{fig:face_rec_chokePt},
it can be observed that using all captured faces generally does not lead to the best performance.
It can also be observed that the proposed method considerably outperforms the other three methods for {\small $N \leq 32$},
and furthermore leads to the best verification performance (which occurs at {\small $N=16$}).
We note that even when only one face is selected by the proposed method (ie.~{\small $N=1$}),
relatively high verification accuracy is still achieved.
This suggests that the proposed method has a high chance of picking the ``best'' face out of a set of faces.

\section{Main Findings}
\label{sec:conclusions}

In this paper we presented a novel patch-based face image quality assessment algorithm.
Unlike previous methods,
the proposed approach is capable of simultaneously handling issues such as 
pose variations, cast shadows, blurriness as well as alignment errors caused by 
automatic face localisation (eg.~in-plane rotations, horizontal and vertical shifts).

The proposed method was
evaluated on two still face datasets (FERET and PIE),
using faces subject to pose and illumination direction changes,
as well as simulated geometric alignment errors and decreased sharpness.
Experiments show that the proposed method has the best overall performance,
identifying images which are the most frontal, well-aligned, illuminated and sharp.
This is accomplished without requiring parameter tuning or retraining for each dataset tested.

The proposed method was also evaluated in a video-based face verification setting,
on a new surveillance dataset termed {\it ChokePoint}.
For each given set of face images for a person,
the proposed method was used to rank the images according to their quality.
By selecting a subset containing only the top quality images,
verification accuracy was considerably improved
when compared to using all available images.
Furthermore, the proposed method consistently led to higher quality subsets
(leading to higher verification accuracy)
than previous image quality assessment algorithms,

The proposed method is capable of assigning low-quality scores to images with cast shadows
(eg.~due to self-shadowing caused by strong directed illumination),
however it is currently unlikely to detect more subtle variations in illumination.
This is due to its elaborate illumination normalisation steps,
necessary for generalisation purposes
(ie.~not being tied to the level of contrast and/or illumination bias in a particular training dataset).
The proposed method is also unlikely to detect minor expression variations,
as only low frequency information is used.
According to~\cite{Aguado10,Swisher04},
expression changes mainly lie in high frequency bands.
However, many of the recent face recognition algorithms
are capable of handling relatively minor variations in both illumination and expression~\cite{Harandi_CVPR_2011,Kumar_ICCV_2009,Phillips_PAMI_10,Sanderson_ICB_2009},
thus these characteristics of the quality assessment method might be more of a feature than a limitation.

\section*{Acknowledgements}

NICTA is funded by the Australian Government
as represented by the {\it Department of Broadband, Communications and the Digital Economy},
as well as the Australian Research Council through the {\it ICT Centre of Excellence} program.
We thank Dr Mehrtash Harandi for useful discussions.
We also thank all the volunteers who participated in the recording of the ChokePoint dataset.

\small
\balance
\bibliographystyle{ieee}
\bibliography{references}

\begin{thebibliography}{10}\itemsep=-1pt

\bibitem{ISO_IEC_19794_5}
\textsc{ISO}/\textsc{IEC} 19794-5 (published version).
\newblock {\em Information technology - Biometric Data Interchange Formats},
  June 2005.

\bibitem{ICAO-9303}
Machine readable travel documents.
\newblock {\em International Civil Aviation Organization}, August 2006.

\bibitem{Aguado10}
L.~Aguado, I.~Serrano-Pedraza, S.~Rodriguez, and F.~J. Roman.
\newblock Effects of spatial frequency content on classification of face gender
  and expression.
\newblock {\em The Spanish Journal of Psychology}, 13(2):525--537, 2010.

\bibitem{Ahonen_ECCV_2004}
T.~Ahonen, A.~Hadid, and M.~Pietik{\"{a}}inen.
\newblock Face recognition with local binary patterns.
\newblock In {\em ECCV, Lecture Notes in Computer Science (LNCS)}, volume 3021,
  pages 469--481, 2004.

\bibitem{Bae_2005}
H.~Bae and S.~Kim.
\newblock Real-time face detection and recognition using hybrid-information
  extracted from face space and facial features.
\newblock {\em Image and Vision Computing}, 23(13):1181--1191, 2005.

\bibitem{Berrani_AVSS_2005}
S.-A. Berrani and C.~Garcia.
\newblock Enhancing face recognition from video sequences using robust
  statistics.
\newblock In {\em IEEE International Conference on Video and Signal Based
  Surveillance (AVSS)}, pages 324--329, 2005.

\bibitem{Chang_CIARP_2008}
L.~Chang, I.~Rod{\'e}s, H.~M{\'e}ndez, and E.~del Toro.
\newblock Best-shot selection for video face recognition using {FPGA}.
\newblock In {\em CIARP, Lecture Notes in Computer Science (LNCS)}, volume
  5197, pages 543--550, 2008.

\bibitem{KANG_EURASIP_2010}
S.~Chen, S.~Mau, M.~T. Harandi, C.~Sanderson, A.~Bigdeli, and B.~C. Lovell.
\newblock Face recognition from still images to video sequences: A
  local-feature-based framework.
\newblock {\em EURASIP Journal on Image and Video Processing}, 2011.

\bibitem{chen_TSMC_2006}
W.~Chen, M.~J. Er, and S.~Wu.
\newblock Illumination compensation and normalization for robust face
  recognition using discrete cosine transform in logarithm domain.
\newblock {\em IEEE Trans. Systems, Man and Cybernetics (Part B)},
  36(2):458--466, 2006.

\bibitem{Gao_ICB_2007}
X.~Gao, S.~Z. Li, R.~Liu, and P.~Zhang.
\newblock Standardization of face image sample quality.
\newblock In {\em ICB, Lecture Notes in Computer Science (LNCS)}, volume 4642,
  pages 242--251, 2007.

\bibitem{Gonzales_2007}
R.~Gonzales and R.~Woods.
\newblock {\em Digital Image Processing}.
\newblock Prentice Hall, 3rd edition, 2007.

\bibitem{modular_PCA}
R.~Gottumukkal and V.~K. Asari.
\newblock An improved face recognition technique based on modular \textsc{PCA}
  approach.
\newblock {\em Pattern Recognition Letters}, 25(4):429--436, 2004.

\bibitem{Hadid_FGR_2004}
A.~Hadid and M.~Pietik{\"a}inen.
\newblock From still image to video-based face recognition: An experimental
  analysis.
\newblock In {\em Proc. Automatic Face and Gesture Recognition (AFGR)}, pages
  813--818, 2004.

\bibitem{Harandi_IJCV_2009}
M.~T. Harandi, M.~N. Ahmadabadi, and B.~N. Araabi.
\newblock {Optimal local basis: A reinforcement learning approach for face
  recognition}.
\newblock {\em International Journal of Computer Vision}, 81(2):191--204, 2009.

\bibitem{Harandi_CVPR_2011}
M.~T. Harandi, C.~Sanderson, S.~Shirazi, and B.~C. Lovell.
\newblock Graph embedding discriminant analysis on {G}rassmannian manifolds for
  improved image set matching.
\newblock In {\em IEEE Conf. Computer Vision and Pattern Recognition (CVPR)},
  pages 2705--2712, 2011.

\bibitem{Hsu_BS_2006}
R.-L.~V. Hsu, J.~Shah, and B.~Martin.
\newblock Quality assessment of facial images.
\newblock In {\em Biometrics Symposium}, 2006.

\bibitem{Kumar_ICCV_2009}
N.~Kumar, A.~Berg, P.~Belhumeur, and S.~Nayar.
\newblock Attribute and simile classifiers for face verification.
\newblock In {\em Int. Conf. Computer Vision (ICCV)}, pages 365--372, 2009.

\bibitem{Luo_icip_2004}
H.~Luo.
\newblock A training-based no-reference image quality assessment algorithm.
\newblock In {\em International Conference on Image Processing (ICIP)}, pages
  2973--2976, 2004.

\bibitem{MOBIO_ICPR_2010}
S.~Marcel, C.~McCool, P.~Matejka, T.~Ahonen, J.~Cernocky, S.~Chakraborty,
  V.~Balasubramanian, S.~Panchanathan, C.~Chan, J.~Kittler, et~al.
\newblock On the results of the first mobile biometry ({MOBIO}) face and
  speaker verification evaluation.
\newblock In {\em Recognizing Patterns in Signals, Speech, Images and Videos,
  Lecture Notes in Computer Science (LNCS)}, volume 6388, pages 210--225, 2010.

\bibitem{Mau_ICVNZ_2010}
S.~Mau, S.~Chen, C.~Sanderson, and B.~C. Lovell.
\newblock Video face matching using subset selection and clustering of
  probabilistic multi-region histograms.
\newblock In {\em International Conference on Image and Vision Computing New
  Zealand (IVCNZ)}, 2010.

\bibitem{Nasrollahi_BIOID_2008}
K.~Nasrollahi and T.~B. Moeslund.
\newblock Face quality assessment system in video sequences.
\newblock In {\em BIOID, Lecture Notes in Computer Science (LNCS)}, volume
  5372, pages 10--18, 2008.

\bibitem{Ozay_CVPRW_2009}
N.~Ozay, Y.~Tong, W.~Frederick~W, and X.~Liu.
\newblock Improving face recognition with a quality-based probabilistic
  framework.
\newblock In {\em Computer Vision and Pattern Recognition (CVPR) Biometrics
  Workshop}, pages 134--141, 2009.

\bibitem{Phillips_PAMI_2000}
P.~J. Phillips, H.~Moon, S.~A. Rizvi, and P.~J. Rauss.
\newblock The {FERET} evaluation methodology for face-recognition algorithms.
\newblock {\em IEEE Trans. Pattern Anal. Mach. Intell.}, 22(10):1090--1104,
  2000.

\bibitem{Phillips_PAMI_10}
P.~J. Phillips, W.~T. Scruggs, A.~J. O'Toole, P.~J. Flynn, K.~W. Bowyer, C.~L.
  Schott, and M.~Sharpe.
\newblock {FRVT} 2006 and {ICE} 2006 large-scale experimental results.
\newblock {\em IEEE Trans. Pattern Anal. Mach. Intell.}, 32(5):831--846, 2010.

\bibitem{Rodriguez_IVC_2006}
Y.~Rodriguez, F.~Cardinaux, S.~Bengio, and J.~Mariethoz.
\newblock Measuring the performance of face localization systems.
\newblock {\em Image and Vision Computing}, 24(8):882--893, 2006.

\bibitem{Rua_BIOID_2008}
E.~A. R{\'u}a, J.~L.~A. Castro, and C.~G. Mateo.
\newblock Quality-based score normalization and frame selection for video-based
  person authentication.
\newblock In {\em BIOID, Lecture Notes in Computer Science (LNCS)}, pages 1--9,
  2008.

\bibitem{Armadillo_2010}
C.~Sanderson.
\newblock Armadillo: An open source {C++} linear algebra library for fast
  prototyping and computationally intensive experiments.
\newblock Technical report, NICTA, 2010.
\newblock \href{http://arma.sourceforge.net/}{\tt
  http://arma.sourceforge.net/}.

\bibitem{Sanderson_ICB_2009}
C.~Sanderson and B.~C. Lovell.
\newblock Multi-region probabilistic \mbox{histograms} for robust and scalable
  identity inference.
\newblock In {\em \mbox{Lecture} Notes in Computer Science (LNCS)}, volume
  5558, pages 199--208, 2009.

\bibitem{ISO-IEC_ICB_2009}
J.~Sang, Z.~Lei, and S.~Z. Li.
\newblock Face image quality evaluation for \textsc{ISO}/\textsc{IEC} standards
  19794-5 and 29794-5.
\newblock In {\em ICB, Lecture Notes in Computer Science (LNCS)}, volume 5558,
  pages 229--238, 2009.

\bibitem{Sanin_ICPR_2010}
A.~Sanin, C.~Sanderson, and B.~C. Lovell.
\newblock Improved shadow removal for robust person tracking in surveillance
  scenarios.
\newblock In {\em International Conference on Pattern Recognition (ICPR)},
  pages 141--144, 2010.

\bibitem{shan_amfg_03}
S.~Shan, W.~Gao, B.~Cao, and D.~Zhao.
\newblock Illumination normalization for robust face recognition against
  varying lighting conditions.
\newblock In {\em Workshop on Analysis and Modeling of Faces and Gestures
  (AMFG)}, pages 157--164, 2003.

\bibitem{Sim_PAMI_2003}
T.~Sim, S.~Baker, and M.~Bsat.
\newblock The {CMU} pose, illumination, and expression database.
\newblock {\em IEEE Transactions on Pattern Analysis and Machine Intelligence},
  25(1):1615 -- 1618, 2003.

\bibitem{Subasic_ISPA_2005}
M.~Subasic, S.~Loncaric, T.~Petkovic, H.~Bogunovic, and V.~Krivec.
\newblock Face image validation system.
\newblock In {\em International Symposium on Image and Signal Processing and
  Analysis (ISPA)}, pages 30--33, 2005.

\bibitem{Swisher04}
J.~D. Swisher, C.~Brooking, and D.~Somers.
\newblock Spatial frequency and facial expressions of emotion.
\newblock {\em Journal of Vision}, 4(8):905, 2004.

\bibitem{low-res-face-detect01}
A.~Torralba and P.~Shina.
\newblock Detecting faces in improverished images.
\newblock {\em Technical Report 028, MIT AI Lab}, 2001.

\bibitem{Wong_ICPR_2010}
Y.~Wong, C.~Sanderson, S.~Mau, and B.~C. Lovell.
\newblock Dynamic amelioration of resolution mismatches for local feature based
  identity inference.
\newblock In {\em International Conference on Pattern Recognition (ICPR)},
  pages 1200--1203, 2010.

\bibitem{lighting_normalization}
X.~Xie and K.-M. Lam.
\newblock An efficient illumination normalization method for face recognition.
\newblock {\em Pattern Recognition Letters}, 27:609--617, 2006.

\bibitem{Yamaguchi_FGR_1998}
O.~Yamaguchi, K.~Fukui, and K.~Maeda.
\newblock Face recognition using temporal image sequence.
\newblock In {\em Proc. Automatic Face and Gesture Recognition (AFGR)}, pages
  318--323, 1998.

\bibitem{Yang_ICPR_2004}
Z.~Yang, H.~Ai, B.~Wu, S.~Lao, and L.~Cai.
\newblock Face pose estimation and its application in video shot selection.
\newblock In {\em International Conference on Pattern Recognition (ICPR)},
  pages 322--325, 2004.

\bibitem{Zhang_isvc_2009}
G.~Zhang and Y.~Wang.
\newblock Asymmetry-based quality assessment of face images.
\newblock In {\em ISVC, Lecture Notes in Computer Science (LNCS)}, volume 5876,
  pages 499--508, 2009.

\bibitem{facerec-survey04}
W.~Zhao, R.~Chellappa, A.~Rosenfeld, and P.~Phillips.
\newblock Face recognition: A literature survey.
\newblock {\em ACM Computing Surveys}, 35(4):399--458, 2003.

\end{thebibliography}

\end{document}